\documentclass[]{template}

\usepackage[utf8]{inputenc}             
\usepackage[T1]{fontenc}                
\usepackage{url}                        
    \usepackage{xurl} 
\usepackage{booktabs}                   
\usepackage{multirow}
\usepackage{colortbl}
\usepackage{multicol}
\usepackage{amsfonts}                   
\usepackage{nicefrac}                   
\usepackage{microtype}                  
\usepackage[dvipsnames]{xcolor}         
\definecolor{qgreen}{RGB}{46, 139, 87}
\providecommand{\hi}{\cellcolor{qgreen!18!white}}

\usepackage{latexsym}

\usepackage{graphicx}
\usepackage{float}
\usepackage{subcaption}
\usepackage{wrapfig}

\usepackage{seqsplit}
\usepackage{etoc}
\AtBeginDocument{\etocdepthtag.toc{main}}

\usepackage{bm}

\usepackage{tabularx} 
\usepackage{ragged2e} 
\newcolumntype{L}{>{\RaggedRight\hangafter=1\hangindent=0em}X}
\newcolumntype{C}{>{\centering\arraybackslash}X}
\newcolumntype{P}[1]{>{\RaggedRight\arraybackslash}p{#1}}
\newcolumntype{Q}[1]{>{\centering\arraybackslash}p{#1}}

\usepackage{enumitem}

\usepackage{amsmath}
\usepackage{amssymb}
\usepackage{mathtools}
\usepackage{amsthm}
\usepackage[linesnumbered,ruled,vlined]{algorithm2e}

\usepackage{hyperref}
\hypersetup{
    colorlinks=true,
    linkcolor=red,
    citecolor=Cerulean,
    filecolor=magenta,      
    urlcolor=magenta,
}

\usepackage[capitalize,noabbrev]{cleveref}
\crefname{section}{§}{§§}
\Crefname{section}{§}{§§}

\usepackage{calligra}
\DeclareMathAlphabet{\mathcalligra}{T1}{calligra}{m}{n}

\usepackage{pifont}

\theoremstyle{plain}

\theoremstyle{definition}

\theoremstyle{remark}

\renewcommand{\paragraph}[1]{\vspace{1mm}\noindent\textbf{#1}}

\DeclareCaptionLabelFormat{cont}{#1~#2\alph{ContinuedFloat}}
\usepackage[most]{tcolorbox}
\tcbset{
  promptbox/.style={
    top=10pt,
    colback=lightgray!20,
    colframe=Black,
    colbacktitle=NavyBlue,
    enhanced,
    center,
    attach boxed title to top center={yshift=-0.1in,xshift=0.0in},
    boxed title style={boxrule=0pt,colframe=white,},
  }
}
\newtcolorbox{promptbox}[2][]{promptbox, title=#2,#1}
\tcbset{
  takeawaybox/.style={
    top=10pt,
    colback=lightgray!20,
    colframe=Black,
    colbacktitle=BurntOrange,
    enhanced,
    center,
    attach boxed title to top center={yshift=-0.1in,xshift=0.0in},
    boxed title style={boxrule=0pt,colframe=white,},
  }
}
\newtcolorbox{takeawaybox}[2][]{takeawaybox, title=#2,#1}
\tcbset{
  observationbox/.style={
    top=10pt,
    colback=lightgray!20,
    colframe=Black,
    colbacktitle=YellowGreen,
    enhanced,
    center,
    attach boxed title to top center={yshift=-0.1in,xshift=0.0in},
    boxed title style={boxrule=0pt,colframe=white,},
  }
}
\newtcolorbox{observationbox}[2][]{observationbox, title=#2,#1}

\usepackage{xspace}

\newcommand\blfootnote[1]{%
  \begingroup
  \renewcommand\thefootnote{}\footnote{#1}%
  \addtocounter{footnote}{-1}%
  \endgroup
}

\title{On-Policy Distillation Teaches New Skills but Not New Knowledge}

\author{Yixuan Tang}
\author{Yi Yang}
\affil{The Hong Kong University of Science and Technology}

\begin{abstract}
\textbf{Abstract: }On-policy distillation (OPD) strengthens language-model reasoning, yet whether students acquire new factual knowledge or compositional skill for multi-step reasoning remains unknown. We separate these capabilities using a controlled synthetic framework that measures the student's initial capabilities and independently controls the teacher's additional facts, compositional skill, or both. Across four models from three families, reverse-KL OPD reliably transfers compositional skill across unseen reasoning structures, but transfers minimal factual knowledge. Decoupling the distillation recipe reveals the source of this asymmetry: replacing reverse KL with forward KL restores factual transfer, whereas student rollouts specifically improve the execution of multi-step reasoning. Experiments on recent factual QA and competition mathematics show a similar asymmetry under reverse-KL OPD, yielding notable reasoning gains without factual memory expansion. Together, these results demonstrate that on-policy distillation does not expand a model's parametric knowledge, but instead teaches it to organize and compose the knowledge it already possesses.
\end{abstract}

\begin{document}
\blfootnote{
Correspondence to: \texttt{ytangch@connect.ust.hk}, \texttt{imyiyang@ust.hk}
}
\maketitle

\section{Introduction}
\label{sec:intro}

Knowledge distillation transfers capabilities from a large teacher model to a smaller student~\citep{hinton2015distilling,kim2016sequence,cloud2026language}. This allows smaller models to approach the performance of capable teachers at lower cost~\citep{zheng2025learning}. Standard distillation, however, trains the student on fixed teacher outputs, so supervision never covers the student's own generation mistakes; this mismatch is known as exposure bias. On-policy distillation (OPD) addresses this limitation by supervising the student along its own generation trajectories~\citep{agarwal2024policy,gu2024minillm,lu2025onpolicydistillation}. Because supervision targets the prefixes the student actually produces, OPD strengthens complex reasoning and is now widely used in post-training pipelines such as Gemini and Qwen~\citep{team2024gemini,2025qwen3}.

Yet these empirical gains leave open what the student learns from the teacher. In multi-step reasoning, reaching an answer requires both recalling domain facts and coordinating intermediate results~\citep{newell1972human}. Higher benchmark scores can therefore reflect newly acquired facts, improved compositional skill, or both, but end-to-end evaluation cannot separate them. These capabilities address distinct computational demands: factual knowledge relies on stored associations in memory, whereas compositional skill determines how intermediate steps are combined~\citep{morris2026how,lake2023human,yuan2026f}. A model can possess the relevant facts yet fail to compose them, or master multi-step reasoning yet fail when querying unfamiliar facts. Understanding what distillation transfers therefore requires evaluating factual knowledge and compositional skill as independent dimensions.

This distinction matters for how on-policy distillation can be applied in practice. If factual knowledge transfers, distillation offers a way to expand a model's parametric memory with new facts. If compositional skill transfers, it teaches the student to use its existing factual knowledge across new reasoning patterns. We therefore ask 
\textit{whether on-policy distillation transfers a teacher's factual knowledge, compositional skill, or both.}


Existing work offers conflicting perspectives on this question. \citet{ding2026does} show that suppressing low-probability student tokens can match OPD's reasoning gains without teacher supervision. In contrast, \citet{li2026rethinking} argue that successful OPD requires the teacher to offer new capabilities beyond what the student has seen during training. Yet neither line of work clarifies whether the student gains new capabilities from the teacher, and whether such gains reflect factual knowledge or compositional skill. Answering these questions requires measuring the student's initial capabilities, controlling the teacher's additional facts and compositional skill, and evaluating both dimensions independently. Such control is difficult with pretrained language models: their training corpora are too vast to audit what they already know, and natural-language tasks often require factual recall and reasoning together.

\begin{figure}[t]
\centering
\includegraphics[width=\textwidth]{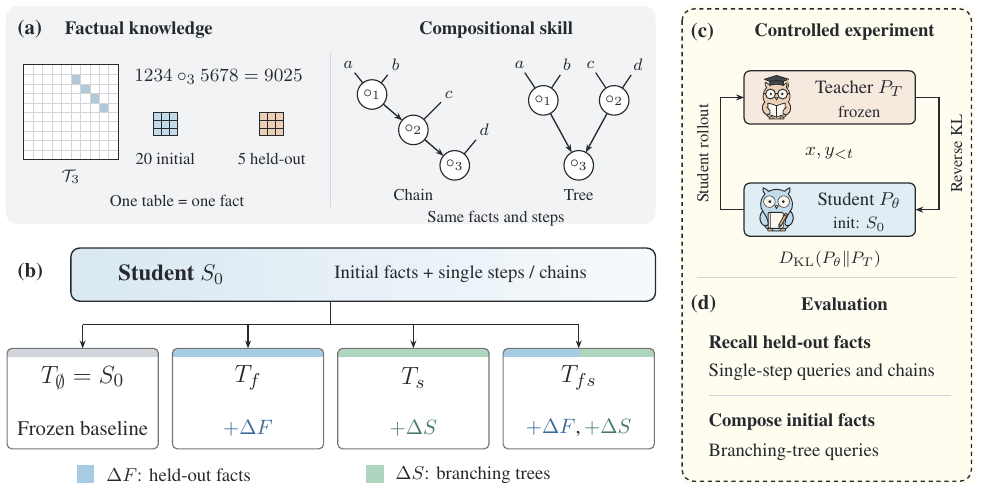}
\caption{Overview of the setup.
(a)~Random lookup tables represent facts; chains and trees specify how their query results are combined.
(b)~Starting from $S_0$, teachers add exposure to held-out facts ($\Delta F$), branching trees ($\Delta S$), or both, with frozen $S_0$ as the baseline.
(c)~A fixed teacher supervises student rollouts with reverse KL.
(d)~Evaluation separates factual recall on held-out facts from compositional skill over initial facts.}
\label{fig:overview}
\end{figure}

To achieve this control, we construct a synthetic environment that explicitly separates factual knowledge from compositional skill (Fig.~\ref{fig:overview}). In natural language, factual knowledge often involves specific associations that cannot be inferred from syntax alone~\citep{morris2026how}. Thus, we operationalize factual knowledge as a set of arbitrary input–output associations stored in a lookup table. Each table maps a pair of input digits to an output digit. To ensure that these associations must be memorized rather than inferred, we sample all table entries independently and uniformly at random, with no underlying algebraic rule. As a result, the answer to an unseen table entry cannot be derived from other entries. The model can answer correctly only if it has learned that specific mapping. This construction also lets us control exactly which facts are available to the student during training.
Beyond recalling individual facts, multi-step reasoning also requires the model to organize and reuse them across successive computations. 
We operationalize compositional skill as the ability to combine facts across multi-step reasoning structures. In a sequential chain, each step directly uses the result of the previous step. In a branching tree, the model must solve separate branches, retain intermediate results across intervening steps, and combine them later in the correct order. By varying the tree structure while holding the underlying facts and number of operations fixed, we test whether the model acquires a general compositional skill rather than memorizing a specific computation pattern.

Within this controlled setup, we establish an initial student, $S_0$, by training it on an initial set of facts using single-step queries and sequential chains. This student reliably recalls the initial facts, but achieves near-zero accuracy on held-out facts and on branching trees. Here, \emph{initial} and \emph{held-out} refer to whether a fact is included during student initialization. 
Starting from \(S_0\), we then train three specialized teachers: \(T_f\) is additionally trained on the held-out facts, \(T_s\) on branching trees using only the initial facts, and \(T_{fs}\) on both. We also retain the unchanged \(S_0\) as a baseline teacher, \(T_{\emptyset}\), which has neither additional knowledge nor additional compositional skill. We then distill each teacher into a student initialized from \(S_0\) and separately evaluate whether the student acquires the teacher’s additional knowledge or compositional skill, including whether the latter generalizes to previously unseen tree structures.

Across four models from three families, our experiments reveal a consistent asymmetry in what is transferred: on-policy distillation with reverse KL transfers compositional skill while failing to transfer factual knowledge. Students distilled from $T_s$ reach 79.10\% accuracy under greedy decoding on unseen tree structure, compared with 0.72\% under the baseline teacher \(T_{\emptyset}\). In contrast, students distilled from $T_f$ reach only 13.12\% on single-step queries about held-out facts and 0.54\% on chains of such facts. To identify the source of this asymmetry, we separate two components of the distillation procedure: the source of training prefixes and the direction of the KL objective. Replacing reverse KL with forward KL substantially restores factual transfer under both student-generated and teacher-generated prefixes, whereas student-generated prefixes primarily improve multi-step compositional execution.
Real-world benchmarks on recent factual question answering and competition mathematics show the same pattern: reverse-KL OPD improves mathematical reasoning by 6.47 percentage points but produces no improvement in factual recall over the initial student.


These findings suggest that reverse-KL based OPD primarily teaches models how to use what they already know, rather than reliably adding what they do not know. 
For practitioners, this distinction suggests that OPD is better suited for strengthening compositional reasoning than for factual knowledge updating, whereas forward-KL objectives are substantially more effective when transferring new factual information is the goal. 


\section{Related Work}


\paragraph{Knowledge Distillation}
Knowledge distillation trains a student to match a teacher's output distribution, allowing a compact model to capture behaviors from a larger counterpart~\citep{hinton2015distilling}. In autoregressive language modeling, sequence-level distillation trains the student with cross-entropy on teacher-generated sequences~\citep{kim2016sequence}. Beyond raw task accuracy, distillation can transfer underlying inductive biases~\citep{ojha2023knowledge} and recover memorized associations that the student never directly observed from a teacher's soft labels~\citep{behrens2026dataset}. However, because sequence-level distillation trains only on teacher-generated sequences, the student receives no supervision along its own generation paths.

\paragraph{On-Policy Distillation}
On-policy distillation addresses this mismatch by evaluating teacher supervision on student-generated trajectories. Generalized Knowledge Distillation (GKD) formalizes this approach across different divergence objectives~\citep{agarwal2024policy}, while MiniLLM uses reverse KL to discourage the student from assigning excessive probability to sequences that are unlikely under the teacher~\citep{gu2024minillm}. Empirical studies report contrasting findings on what students acquire through this supervision. Some studies report evidence of capability expansion: the Qwen3 technical report shows that on-policy distillation improves Pass@64 on math reasoning, whereas reinforcement learning from the same checkpoint leaves Pass@64 unchanged~\citep{2025qwen3}. In addition, SDFT demonstrates factual knowledge acquisition through on-policy self-distillation from a teacher conditioned on additional information~\citep{shenfeld2026selfdistillation}. In contrast, other studies argue that reasoning gains mainly reflect more efficient sampling of existing capabilities. At larger sampling budgets, pre-OPD models can match or outperform their distilled counterparts, suggesting improved access to existing reasoning capabilities rather than a broader capability boundary~\citep{ge2026towards,wang2026demystifying}. Furthermore, \citet{ding2026does} show that suppressing low-probability student tokens without teacher supervision can match OPD's reasoning gains. These findings concern different sources of teacher information and different measures of student capability. To address these confounding factors, we independently control the additional knowledge and compositional skill available to the teacher and evaluate whether each is transferred to the student. 

\paragraph{Synthetic Tasks for Factual Knowledge and Compositional Skill}
Controlled synthetic tasks make it possible to precisely control training exposure and measure generalization. \citet{morris2026how} measure memorization capacity using uniformly random sequences with known information content. To study compositional skill, \citet{yuan2026f} construct string-transformation tasks in which models first learn atomic functions and then learn to compose them through reinforcement learning. Synthetic tasks have also been used to study distillation: \citet{behrens2026dataset} show that teacher soft labels can convey memorized associations that students never directly observe. We use controlled synthetic tasks to examine factual knowledge and compositional skill transfer under on-policy distillation. Each random lookup table represents a piece of factual knowledge, while chains and branching trees determine how known facts must be combined during multi-step reasoning. By independently varying the additional knowledge and skill available to the teacher, we test what the student acquires through distillation.

\section{Experimental Setting: Separating Knowledge and Skill}
To identify what is transferred during on-policy distillation, we need to control both the student’s initial capabilities and the teacher’s additional knowledge and skill. We therefore construct a controlled setting in which these capabilities can be varied independently.


\subsection{Synthetic Tasks for Knowledge and Reasoning}
\label{sec:synthetic_world}

\paragraph{Factual Knowledge as Random Lookup Tables}
Isolating factual transfer requires distinguishing memorization from rule-based generalization~\citep{morris2026how}. If a task contains learnable patterns, a model can deduce unseen answers rather than recall the specified mapping. We therefore operationalize each fact as a two-argument mapping represented by one random lookup table.  Because the entries have no underlying algebraic structure, the model must learn these associations to answer queries about the fact. Therefore, withholding selected tables during student initialization controls which facts the student encounters.

We instantiate each fact as a lookup table $\mathcal{T}_k:\{0,\ldots,9\}^2\to\{0,\ldots,9\}$. Each table contains 100 entries, each mapping an input digit pair to an output digit. We use $a\circ_k b$ to denote the result of applying table $\mathcal{T}_k$ independently at each digit position:
\begin{equation}
    (a \circ_k b)_j = \mathcal{T}_k(a_j, b_j), \qquad j\in\{1,2,3,4\}.
    \label{eq:digitwise_operation}
\end{equation}
A single-step fact query therefore requires four digit lookups with no carry across positions. For example, fact $\mathcal{T}_3$ is identified  by \texttt{[OP\_3]}, and the query \texttt{1234[OP\_3]5678} evaluates to \texttt{9025}. The lookup table itself is not included in the prompt, so answering correctly requires the corresponding associations to have been learned during training. We generate 25 facts, $\mathcal{T}=\{\mathcal{T}_1,\ldots,\mathcal{T}_{25}\}$, partitioned into 20 \emph{initial facts}, $\mathcal{T}_{\mathrm{init}}$, and five \emph{held-out facts}, $\mathcal{T}_{\mathrm{held}}$. Initial facts are included during student initialization, whereas held-out facts are excluded. This partition remains fixed throughout teacher training and distillation. The tokens \texttt{[OP\_1]} through \texttt{[OP\_25]} identify the facts in the serialized input and output.

\paragraph{Compositional Skill across Reasoning Structures}
Compositional skill is the ability to order fact queries and supply intermediate results according to a computation graph~\citep{newell1972human,yuan2026f}. To isolate this ability from factual recall, we construct sequential chains and branching trees that query the same set of facts and require the same number of operations. In a chain, each step simply takes the result of the immediately preceding step and applies the next fact. 
In a branching tree, independent branches are evaluated before their results are combined. Although the output is still generated sequentially, the model must retrieve earlier intermediate results across intervening steps to combine independent branches. In our setting, chains are used during student initialization, while branching trees test compositional skill without introducing new facts.

A simple three-step example illustrates how intermediate results are passed in each structure:
\[
\begin{array}{ll}
\text{Chain: }((a\circ_1 b)\circ_2 c)\circ_3 d
& \text{Tree: }(a\circ_1 b)\circ_3(c\circ_2 d) \\[3pt]
r_1=a\circ_1 b & r_1=a\circ_1 b \\
r_2=r_1\circ_2 c & r_2=c\circ_2 d \\
r_3=r_2\circ_3 d & r_3=r_1\circ_3 r_2.
\end{array}
\]
Here, $a, b, c, d$ are four-digit inputs, and $\circ_i$ denotes a query to fact $\mathcal{T}_i$. In the chain, step 3 uses the immediately preceding result $r_2$ and input $d$. In the tree, step 3 combines $r_1$ and $r_2$, requiring retrieval of $r_1$ across the intervening computation of $r_2$. 
Thus, the two structures use the same facts but differ in how intermediate results must be organized and reused. In our main experiments, each query contains seven operations over eight inputs. Each completion includes an explicit step-by-step execution trace followed by the final answer. We measure the student's initial performance on both chain and branching-tree structures in Section~\ref{sec:student_pretraining}. Appendix~\ref{sec:execution_examples} provides complete prompt and completion examples.



\subsection{Student Initialization and Teacher Design}
We establish an initial student, \(S_0\), by training it on an initial set of facts and sequential chains.  We then verify that it reliably recalls these facts and executes chains, but performs poorly on held-out facts and branching trees.

\paragraph{Initial Student}
\label{sec:student_pretraining}
We use four base models: Qwen3-0.6B and Qwen3-4B~\citep{2025qwen3}, Gemma-2-2B~\citep{team2024gemma}, and OLMo3-7B~\citep{olmo2025olmo}. All distillation experiments begin from the initial student, $S_0$. We obtain $S_0$ by fine-tuning a base LLM on queries about the initial facts, $\mathcal{T}_{\mathrm{init}}$, using single-step queries and sequential chains of varying lengths. This initialization teaches the output format and factual recall while excluding data about held-out facts, $\mathcal{T}_{\mathrm{held}}$, and branching trees. Evaluation confirms that $S_0$ reliably executes chains over initial facts but rarely solves branching trees or queries about held-out facts. Appendix~\ref{sec:v31_details} reports the training mixture and baseline measurements.

\paragraph{Controlled Teacher Design}
\label{sec:teacher_design}
Starting from $S_0$, we train three specialized teachers by augmenting its training distribution with data about held-out facts ($T_f$), branching trees over initial facts ($T_s$), or both ($T_{fs}$), corresponding to $\Delta F$ and $\Delta S$ in Fig.~\ref{fig:overview}(b). Together with the frozen baseline teacher $T_\emptyset=S_0$, they form a $2\times2$ factorial design crossing exposure to held-out facts with exposure to branching trees. Here, $f$ denotes factual knowledge and $s$ denotes compositional skill. All teachers share the student's base architecture and initialization. Table~\ref{tab:training_conditions}(a) summarizes the factual knowledge and compositional skill of the initial student and teachers. Evaluations confirm the intended capability separation. For example, using Qwen3-4B as base model,  \(T_f\) reaches 100\% accuracy on held-out factual knowledge queries but only 0.60\% on unseen branching trees, whereas \(T_s\) reaches 100\% on unseen trees but 0\% on held-out factual queries (Table~\ref{tab:teacher_profiles_eval}).

\begin{table}[htbp]
\centering
\captionsetup{justification=centering,singlelinecheck=true}
\caption{Student and teacher capabilities and distillation conditions.}
\label{tab:training_conditions}
\small
\setlength{\tabcolsep}{9pt}
\renewcommand{\arraystretch}{1.15}
\begin{tabular}{@{}lll@{}}
\multicolumn{3}{@{}l}{\textbf{(a) Student and teacher capabilities}} \\[3pt]
\toprule
Model & Factual knowledge & Compositional skill \\
\midrule
$S_0$ & Initial facts & Chain execution \\
$T_\emptyset=S_0$ & Initial facts & Chain execution \\
$T_f$ & Initial + held-out facts & Chain execution \\
$T_s$ & Initial facts & Chain and branching-tree execution \\
$T_{fs}$ & Initial + held-out facts & Chain and branching-tree execution \\
\bottomrule
\addlinespace[10pt]
\multicolumn{3}{@{}l}{\textbf{(b) On-policy distillation}} \\[3pt]
\toprule
Fixed teacher & Student starts from & Distillation prompts \\
\midrule
$T_f$ & $S_0$ & Chains querying held-out facts \\
$T_s$ & $S_0$ & Branching trees over initial facts \\
$T_{fs}$ & $S_0$ & Branching trees querying held-out facts \\
$T_\emptyset$ & $S_0$ & Same queries as the $T_f$ run \\
$T_\emptyset$ & $S_0$ & Same queries as the $T_s$ run \\
\bottomrule
\end{tabular}
\end{table}

\subsection{On-Policy Distillation and Evaluation}
\label{sec:evaluation_protocol}

\paragraph{Training Objective}
\label{sec:distillation_framework}
All distillation runs start from the same student $S_0$ and keep the selected teacher $T$ fixed. Given a prompt $x$, the current student policy $P_\theta$ samples a completion $y\sim P_\theta(\cdot\mid x)$, which is called a rollout. At each position $t$, the teacher's next-token distribution $P_T$ is conditioned on the same prompt and the student's preceding tokens $y_{<t}$. We train the student to match this distribution using token-level reverse KL~\citep{agarwal2024policy,gu2024minillm}. For a batch $\mathcal{B}$ of prompt-completion pairs, the loss is
\begin{equation}
    \widehat{\mathcal{L}}_{\mathrm{OPD}}(\theta;\mathcal{B})
    =\frac{1}{Z_{\mathcal{B}}}
    \sum_{(x,y)\in\mathcal{B}}\sum_{t=1}^{|y|}
    D_{\mathrm{KL}}\!\left(
    P_\theta(\cdot\mid x,y_{<t})\,\Vert\,P_T(\cdot\mid x,y_{<t})
    \right),
    \label{eq:opd_objective}
\end{equation}
where $Z_{\mathcal{B}}$ counts completion tokens, excluding prompt and padding positions. Sampled completions are held fixed during each update, and gradients pass only through $P_\theta$. Optimization details appear in Appendix~\ref{sec:v31_details}.

\paragraph{Distillation Conditions}
To test factual knowledge transfer, we distill from \(T_f\) using multi-step chains that contain held-out facts. To test compositional skill transfer, we distill from \(T_s\) using multi-step trees over initial facts. The joint condition uses \(T_{fs}\) with trees that also contain held-out facts. As a baseline, we  use $T_\emptyset=S_0$ on the same prompts to provide a control with no additional teacher knowledge or skill. Dataset sizes and mixtures appear in Appendix~\ref{sec:training_corpora}.

\paragraph{Evaluation}
We evaluate factual recall with single-step queries and multi-step chains over held-out facts. Compositional skill is evaluated on multi-step trees over initial facts. To measure generalization, we divide tree structures into three groups:  \emph{Shared} structures appear in both teacher training and distillation. \emph{Teacher-only} structures appear only in teacher training. \emph{Unseen} structures appear in neither. 
Distillation from \(T_s\) and \(T_{fs}\) uses only shared structures, while evaluation covers all three groups using fresh queries.
Appendix~\ref{sec:tests_metrics} provides the split details, test definitions, and metrics.

\section{Synthetic Experiment Results}
\subsection{On-Policy Distillation Teaches New Skills but Not New Knowledge}

\begin{figure}[t]
\centering
\includegraphics[width=\textwidth]{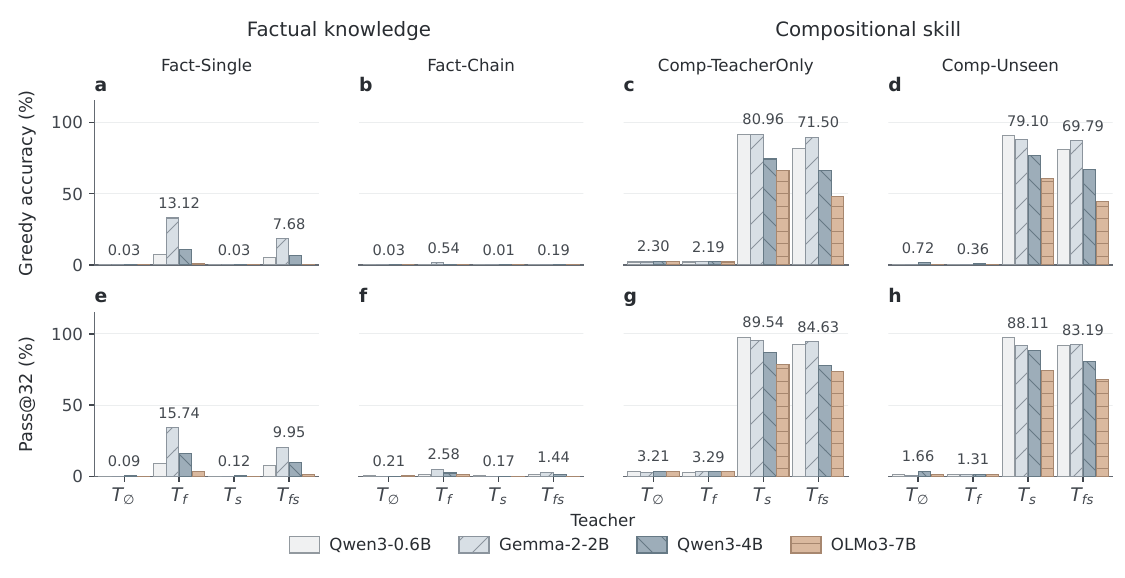}
\caption{Student performance after on-policy distillation under each teacher. Top: greedy final-answer accuracy; bottom: final-answer Pass@32 (at least one of 32 sampled completions correct). Numbers above each group show the unweighted mean across four models. All values are percentages.}
\label{fig:transfer_by_axis}
\end{figure}

Figure~\ref{fig:transfer_by_axis} shows that on-policy distillation transfers compositional skill beyond the tree structures used during training. Averaged across the four models, students distilled from $T_s$ achieve 80.96\% greedy accuracy on teacher-only tree structures and 79.10\% on unseen structures, compared with just 0.72\% on the same unseen structures under the matched $T_\emptyset$ control. The high accuracy on unseen tree structures shows that the acquired compositional skill generalizes beyond the structures encountered during training. Importantly, both the teacher-only and unseen tests use only facts that \(S_0\) had already learned during initialization. The improvement therefore reflects a new ability  to organize and combine existing knowledge.

Factual knowledge transfer is far more limited and varies across models. Students distilled from $T_f$ achieve a mean greedy accuracy of 13.12\% on single-step queries about held-out facts, compared with 0.03\% under the matched $T_\emptyset$ control. Gemma-2-2B reaches 32.95\%, showing that factual recall can improve in certain architectures. However, accuracy drops to just 0.54\% on multi-step chains where every step queries a held-out fact.

This asymmetry between factual knowledge and compositional skill transfer persists under a larger sampling budget. Pass@32 counts a query as solved when at least one of 32 sampled completions has the correct final answer. Students distilled from $T_s$ achieve a mean Pass@32 of 88.11\% on unseen tree structures. In contrast, students distilled from $T_f$ reach only 15.74\% Pass@32 on single-step queries and 2.58\% on chains, both querying held-out facts. The joint $T_{fs}$ condition exhibits the same disparity: students acquire robust compositional skills across tree structures while  unable to reliably recall or chain new  facts. Rollout accuracy and reverse-KL training curves are reported in Appendix~\ref{sec:training_dynamics} (Figure~\ref{fig:training_curves}).

\begin{observationbox}{Insight 1: Transfer Asymmetry in Standard OPD}
On-policy distillation transfers compositional skill, but fails to transfer factual knowledge.
\end{observationbox}

\subsection{Distilling the Distillation Recipe}
\label{sec:distillation_recipe}

\begin{figure}[t]
\centering
\includegraphics[width=\textwidth]{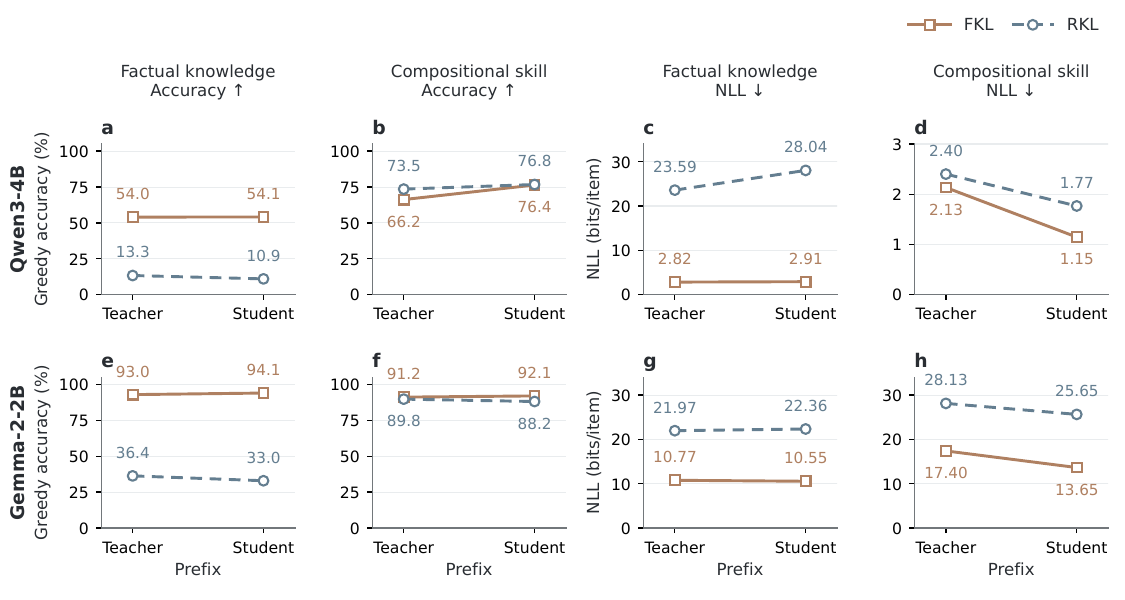}
\caption{Effects of prefix source and KL direction. Rows show Qwen3-4B and Gemma-2-2B. Factual knowledge is evaluated after distillation from $T_f$; compositional skill after distillation from $T_s$. Accuracy measures greedy final answers; NLL is reported for reference completions in bits per completion. Reference prefixes come from teacher demonstrations.}
\label{fig:distillation_recipe}
\end{figure}

Why does on-policy distillation transfer compositional skill but fail on factual knowledge? In sequence distillation, the training procedure combines two independent design choices~\citep{czarnecki2019distilling,agarwal2024policy}: \emph{where the training prefixes come from}, and \emph{how the student matches the teacher's next-token distribution}. Standard OPD couples student rollouts with reverse KL, whereas offline distillation couples teacher demonstration prefixes with forward KL. We therefore separate these two factors to determine which one drives the observed transfer asymmetry. 

The first choice is the \emph{prefix source}. To predict token $t$, the student conditions on a prefix of earlier tokens $y_{<t}$. In offline teacher-forcing, this prefix is drawn from ground-truth teacher demonstrations ($y_{<t} = y^*_{<t}$), training the student exclusively along correct execution paths. In on-policy distillation, the student generates its own prefixes through autoregressive rollouts ($y_{<t} \sim P_\theta$), extending supervision to intermediate states that follow its own generation mistakes.

The second choice is the \emph{divergence objective}. At each prefix $y_{<t}$, the frozen teacher supplies a target next-token distribution $P_T(\cdot \mid y_{<t})$, and the student predicts $P_\theta(\cdot \mid y_{<t})$. Standard OPD minimizes reverse KL, $D_{\mathrm{KL}}(P_\theta\Vert P_T)$~\citep{gu2024minillm,lu2025onpolicydistillation}, which is mode-seeking and weights prediction errors by the student's own probabilities. In contrast, offline distillation typically minimizes forward KL, $D_{\mathrm{KL}}(P_T\Vert P_\theta)$, which is mean-covering and penalizes the student whenever it assigns low probability to tokens favored by the teacher.

\paragraph{Evaluation}
Crossing these two choices yields a $2\times2$ factorial design. We evaluate all four combinations on Qwen3-4B and Gemma-2-2B. We test factual recall on single-step queries about held-out facts after distillation from $T_f$, and compositional skill on unseen tree structures after distillation from $T_s$. All runs share the initial checkpoint $S_0$, training data, and optimization hyperparameters. We also report negative log-likelihood on teacher demonstrations (NLL, in bits per completion) to measure the probability mass the student assigns to correct multi-step executions independently of greedy decoding choices.

\paragraph{Factual Transfer Depends on the Divergence Objective}
Figure~\ref{fig:distillation_recipe} shows that factual recall depends almost entirely on the divergence objective, not the prefix source. Under student prefixes, replacing reverse KL with forward KL substantially increases single-step greedy accuracy: from 10.95\% to 54.10\% in Qwen3-4B, and from 32.95\% to 94.10\% in Gemma-2-2B. Under teacher prefixes, forward KL reaches similar accuracy (54.05\% in Qwen3-4B and 93.00\% in Gemma-2-2B), showing that on-policy rollouts provide no distinct advantage for learning new facts. Reference-completion NLL confirms this pattern, dropping substantially under forward KL across both prefix sources.

These results show that OPD's failure on factual knowledge stems from the divergence objective, not from on-policy exploration. Because reverse KL evaluates discrepancies under the student's distribution, it only penalizes mistakes on tokens the student already produces with substantial probability. When the student does not know a fact, it assigns negligible probability to the correct target, incurring almost no penalty for the omission. Forward KL, by contrast, evaluates discrepancies under the teacher's distribution, heavily penalizing the student whenever it misses the teacher's target.

\paragraph{Student Rollouts Strengthen Compositional Skill}
Student rollouts strengthen compositional skill on unseen tree structures. Under both KL directions, training on student prefixes yields lower reference-completion NLL in both models. Greedy accuracy also improves in three of the four comparisons. For Qwen3-4B, switching from teacher to student prefixes reduces NLL from 2.40 to 1.77 bits under reverse KL and from 2.13 to 1.15 bits under forward KL. For Gemma-2-2B, the corresponding reductions are from 28.13 to 25.65 bits and from 17.40 to 13.65 bits. Thus, training on the student's own trajectories improves the average log-likelihood of correct reference completions, even when the evaluated tree structures were absent from training.

\begin{observationbox}{Insight 2: Decoupling the Distillation Recipe}
The KL direction determines factual knowledge transfer, while on-policy exploration drives compositional skill transfer.
\end{observationbox}

\section{Real-World Validation}
\label{sec:real_world}

To test whether the transfer asymmetry observed in synthetic tasks holds in natural language settings, we evaluate on-policy distillation on recent factual knowledge acquisition and mathematical reasoning. These domains represent two different capabilities: 
factual question answering requires acquiring information that the student does not initially know, whereas mathematical reasoning requires applying and composing existing knowledge across multiple steps.

\paragraph{Benchmark and Teacher Design}
We conduct experiments on Qwen3-4B-Base. We first fine-tune this model on general-instruction data from Tulu-3-SFT~\citep{lambert2024tulu} to establish our initial student $S_0$. We curate two complementary evaluation suites:
\begin{itemize}[leftmargin=*,noitemsep,topsep=2pt]
    \item For recent factual knowledge, \textsc{Knowledge-2026} aggregates 809 question-answer pairs that entered public discourse in 2026, combining items from RealTimeQA~\citep{kasai2023realtime}, SealQA~\citep{pham2026sealqa}, FreshQA~\citep{vu2024freshllms}, and newly edited Wikipedia articles extracted via the WikiBigEdit pipeline~\citep{pmlr-v267-thede25a}. Because these items postdate the training cutoff of Qwen3-4B-Base, $S_0$ achieves only 11.25\% greedy accuracy.
    \item For mathematical reasoning, \textsc{Math-2026} contains 201 competition problems drawn from MathArena~\citep{balunovic2026matharena} and NuminaMath-CoT~\citep{numina_math_datasets}, where $S_0$ achieves 17.41\% greedy accuracy. The 201 problems are held out from a curated pool of 1,000. The remaining 799 serve as distillation prompts.
\end{itemize}

Starting from $S_0$, we train two specialized teachers by supervised fine-tuning. The knowledge teacher $T_K$ is fine-tuned on the full set of 809 factual question-answer pairs, and its 84.55\% on \textsc{Knowledge-2026} is measured on this same set. The reasoning teacher $T_R$ is fine-tuned on competition mathematics with verified step-by-step solutions~\citep{du2026nemotronmath}, reaching 43.78\% on \textsc{Math-2026} (Table~\ref{tab:real-opd}). Further benchmark and teacher details appear in Appendix~\ref{sec:real_world_details}.

\paragraph{On-Policy Distillation Protocol and Results}
To test factual transfer, we distill $S_0$ from $T_K$ using these same factual question prompts for student rollouts. To test reasoning transfer, we distill $S_0$ from $T_R$ over these 799 problem prompts without demonstration traces. Table~\ref{tab:real-opd} compares performance before and after distillation.

\begin{table}[htbp]
  \centering
  \caption{Greedy accuracy (\%) before and after on-policy distillation  on real-world benchmarks. Shaded cells indicate each teacher's target capability.}
  \label{tab:real-opd}
  \footnotesize
  \setlength{\tabcolsep}{10pt}
  \renewcommand{\arraystretch}{1.15}
  \begin{tabular}{@{}llcc@{}}
    \toprule
    \textbf{Model} & \textbf{Role / Condition} & \textbf{\textsc{Knowledge-2026}} & \textbf{\textsc{Math-2026}} \\
    \midrule
    \multirow{5}{*}{\textbf{Qwen3-4B}}
     & Initial student $S_0$ & 11.25 & 17.41 \\
     \cmidrule(lr){2-4}
     & Knowledge teacher $T_K$ & \hi 84.55 & 13.93 \\
     & Reasoning teacher $T_R$ & 12.24 & \hi 43.78 \\
     \cmidrule(lr){2-4}
     & Distilled from $T_K$ & 9.64 & 19.40 \\
     & Distilled from $T_R$ & 10.75 & 23.88 \\
    \bottomrule
  \end{tabular}
\end{table}

The real-world outcomes closely mirror our findings in the synthetic setting. For recent factual knowledge, on-policy distillation fails to transfer the teacher's knowledge: student accuracy drops slightly from 11.25\% to 9.64\%, remaining near its initial level despite the teacher reaching 84.55\%. Thus, even when the student is explicitly prompted with missing facts during distillation, factual recall remains near its initial level.

For mathematical reasoning, on-policy distillation successfully transfers the teacher's multi-step capability: distillation from $T_R$ increases greedy accuracy on \textsc{Math-2026} from 17.41\% to 23.88\% (a 6.47 percentage point gain). This acquired skill generalizes effectively to competition problems excluded from distillation. Neither distillation run produces cross-capability transfer: distilling from $T_R$ leaves factual recall at 10.75\% (matching $S_0$'s 11.25\%), while distilling from $T_K$ shifts math accuracy only to 19.40\%. Therefore, real-world tasks show the same transfer asymmetry: standard OPD improves reasoning skills but transfers factual knowledge only weakly.

\section{Additional Analysis}

\subsection{Execution Traces of Factual and Compositional Failures}
\label{sec:trace_analysis}

\begin{figure}[t]
\centering
\begin{tcolorbox}[colback=lightgray!12,colframe=Black,boxrule=0.4pt,left=8pt,right=8pt,top=8pt,bottom=8pt]
\footnotesize
\textbf{(a)} \textit{Chain querying only held-out facts: correct structure, wrong values}\par\smallskip
{\ttfamily\noindent
((((((3486 [OP\_24] 2845) [OP\_24] 0536) [OP\_22] 7919) [OP\_24] 1381)\newline
\hspace*{1em}[OP\_22] 2146) [OP\_25] 4658) [OP\_25] 3844 = ?\par}
\smallskip
\textit{Student distilled from $T_f$ with reverse KL}\par\smallskip
{\ttfamily\noindent
3486[OP\_24]2845={\color{red}0135};\;
0135[OP\_24]0536={\color{red}0426};\;
0426[OP\_22]7919={\color{red}5020};\;\newline
5020[OP\_24]1381={\color{red}3609};\;
3609[OP\_22]2146={\color{red}1948};\;
1948[OP\_25]4658={\color{red}1332};\;\newline
1332[OP\_25]3844={\color{red}2752};\;
[ANS] {\color{red}2752} [/ANS]\par}
\smallskip
\textit{Student distilled from $T_f$ with forward KL (matches teacher demonstration)}\par\smallskip
{\ttfamily\noindent
3486[OP\_24]2845=8135;\;
8135[OP\_24]0536=9422;\;
9422[OP\_22]7919=0726;\;\newline
0726[OP\_24]1381=3007;\;
3007[OP\_22]2146=1326;\;
1326[OP\_25]4658=3440;\;\newline
3440[OP\_25]3844=5149;\;
[ANS] {\color{qgreen}5149} [/ANS]\par}
\end{tcolorbox}

\medskip
\begin{tcolorbox}[colback=lightgray!12,colframe=Black,boxrule=0.4pt,left=8pt,right=8pt,top=8pt,bottom=8pt]
\footnotesize
\textbf{(b)} \textit{Unseen tree over initial facts: correct values, flattened structure}\par\smallskip
{\ttfamily\noindent
((2250 [OP\_8] 0227) [OP\_6] (3754 [OP\_20] 0476)) [OP\_16]\newline
\hspace*{1em}(((9734 [OP\_1] 6055) [OP\_12] 7785) [OP\_4] 8848) = ?\par}
\smallskip
\textit{Student distilled from $T_f$ with forward KL}\par\smallskip
{\ttfamily\noindent
2250[OP\_8]0227=8071;\;
8071[OP\_6]{\color{red}3754}=6519;\;
6519[OP\_20]0476=7364;\;\newline
9734[OP\_1]6055=0619;\;
0619[OP\_12]7785=1563;\;
1563[OP\_4]8848=7255;\;\newline
{\color{red}7255[OP\_16]7364=8812;}\;
[ANS] {\color{red}8812} [/ANS]\par}
\smallskip
\textit{Student distilled from $T_s$ with reverse KL (matches teacher demonstration)}\par\smallskip
{\ttfamily\noindent
2250[OP\_8]0227=8071;\;
3754[OP\_20]0476=8437;\;
8071[OP\_6]8437=0493;\;\newline
9734[OP\_1]6055=0619;\;
0619[OP\_12]7785=1563;\;
1563[OP\_4]8848=7255;\;\newline
0493[OP\_16]7255=2271;\;
[ANS] {\color{qgreen}2271} [/ANS]\par}
\end{tcolorbox}
\caption{Greedy execution traces from Qwen3-4B students after distillation. (a)~Seven-step chain over held-out facts: reverse KL maintains correct execution syntax but outputs incorrect lookup values (\textcolor{red}{red}), whereas forward KL restores correct lookups. (b)~Unseen tree over initial facts: forward KL recalls all facts correctly but flattens the tree into a linear sweep (\textcolor{red}{red}), whereas reverse KL from $T_s$ executes the valid postorder. Red highlights incorrect values or misrouted operands; green marks correct final answers.}
\label{fig:trace_mirror}
\end{figure}

To examine how distillation failures manifest during generation, we inspect greedy execution traces from Qwen3-4B on held-out chains and unseen trees (Figure~\ref{fig:trace_mirror}).

In Figure~\ref{fig:trace_mirror}(a), the student is distilled from $T_f$ with reverse KL and evaluated on a chain over held-out facts. The student follows the step-by-step format without error, passing intermediate outputs to the next query and closing with an answer tag. What fails is the lookup facts: each step outputs an incorrect number from the queried table, such as \texttt{0135} instead of \texttt{8135}. Across evaluation chains, the execution format never breaks, but most individual lookups are incorrect. Distillation from the same teacher with forward KL provides the target entries and matches the teacher demonstration.

In Figure~\ref{fig:trace_mirror}(b), the student is distilled from $T_f$ with forward KL and evaluated on an unseen tree over initial facts. Here, the student recalls every queried fact correctly, but flattens the tree into a left-to-right chain. After evaluating the first pair of leaves, it takes the next number \texttt{3754} as an operand rather than resolving the subtree \texttt{3754 [OP\_20] 0476} first. Because $T_f$ does not have tree skill, the student never learns how to navigate branching dependencies during its own rollouts. Distillation from $T_s$ with reverse KL corrects the evaluation order and solves the tree. In summary, these examples show that knowledge failures produce wrong values, whereas skill failures produce wrong reasoning order.

\subsection{Weight Updates under Factual and Compositional Distillation}
\label{sec:adapter_anatomy}

We next examine how factual and compositional distillation reshape the student's parameters by analyzing the merged LoRA updates relative to $S_0$ in Qwen3-4B, Gemma-2-2B, and OLMo3-7B (Table~\ref{tab:adapter_anatomy}). Across all three models, two clear mechanistic patterns emerge.

First, factual and compositional distillation occupy near-orthogonal parameter subspaces. Across matched prefix sources and divergence objectives, the cosine similarity between updates from $T_f$ and $T_s$ is near zero, ranging from 0.004 to 0.046. In contrast, changing the divergence objective while holding the task fixed preserves substantial directional alignment, reaching cosine similarities of 0.64--0.82 under $T_s$. This demonstrates that storing facts and acquiring compositional skills recruit distinct directions in parameter space rather than competing for the same weights. 

Second, the two capabilities modify distinct network depths. In Qwen3-4B, updates from $T_f$ concentrate in the final layers, increasing monotonically with depth and peaking at layer 35 where final vocabulary predictions are formed. Conversely, updates from $T_s$ concentrate in the early and middle layers, peaking at layers 3 and 11 before decaying toward the output. This divergence aligns with functional roles in Transformer architectures, where late layers mediate token-level factual associations while earlier layers route and assemble intermediate representations across multi-step dependencies. Appendix~\ref{sec:adapter_metrics} provides full breakdowns across individual projections, singular value distributions, and embedding drift (Table~\ref{tab:adapter_anatomy_full}).

\begin{observationbox}{Insight 3: Orthogonal Subspaces and Depth Specialization}
Factual and compositional updates occupy near-orthogonal parameter directions and separate across network depths, with facts concentrating in final layers and compositional skill reshaping early routing layers.
\end{observationbox}

\begin{table}[htbp]
\centering
\small
\caption{LoRA weight updates relative to $S_0$. (a) Update magnitude and directional alignment across models. (b) Layer-wise norm distribution across depth quartiles in Qwen3-4B. Full breakdowns across all projections and embedding drift appear in Appendix~\ref{sec:adapter_metrics} (Table~\ref{tab:adapter_anatomy_full}).}
\label{tab:adapter_anatomy}
\setlength{\tabcolsep}{8pt}
\renewcommand{\arraystretch}{1.08}
\begin{tabular}{@{}lccc@{}}
\toprule
\textbf{(a) Parameter Metric} & Qwen3-4B & Gemma-2-2B & OLMo3-7B \\
\midrule
Update Norm: $T_f$ (factual) & 2.01--2.86 & 1.33--1.52 & 3.08--3.89 \\
Update Norm: $T_s$ (compositional) & 1.27--1.55 & 1.13--1.60 & 2.50--3.21 \\
Update Norm: $T_\emptyset$ (control) & 0.30--0.32 & 0.20--0.24 & 0.48--0.55 \\
\midrule
Cosine Similarity: $T_f$ vs. $T_s$ & 0.004--0.018 & 0.011--0.036 & 0.004--0.046 \\
Cosine Similarity: Rev. vs. Fwd. KL ($T_s$) & 0.73--0.78 & 0.64--0.82 & 0.68--0.80 \\
\bottomrule
\end{tabular}
\par\smallskip
\begin{tabular}{@{}lccccc@{}}
\toprule
\textbf{(b) Qwen3-4B Layer Norms} & L0--8 & L9--17 & L18--26 & L27--35 & Peak Layer \\
\midrule
$T_f$ (factual knowledge) & 0.24--0.33 & 0.25--0.39 & 0.33--0.46 & 0.46--0.65 & Layer 35 \\
$T_s$ (compositional skill) & 0.25--0.29 & 0.26--0.27 & 0.18--0.25 & 0.11--0.24 & Layers 3, 11 \\
$T_\emptyset$ (control baseline) & 0.04--0.05 & 0.05--0.05 & 0.05--0.05 & 0.06--0.06 & Layer 29 \\
\bottomrule
\end{tabular}
\end{table}

\subsection{Transfer under Alternative Distillation Objectives}
\label{sec:opd_variants}

We extend the comparison in Section~\ref{sec:distillation_recipe} to objectives that combine the effects of forward and reverse KL, keeping student rollouts as the prefix source. On Qwen3-4B, we evaluate GKD with Jensen--Shannon divergence ($\beta=0.5$)~\citep{agarwal2024policy}, ToDi with a sigmoid-weighted combination of forward and reverse KL for each vocabulary token~\citep{jung2025todi}, and EOPD, which adds forward KL to a reverse-KL objective at positions with high teacher entropy~\citep{jin2026entropyaware}. Each objective is evaluated with $T_f$ and $T_s$, starting from $S_0$.

\begin{table}[htbp]
\centering
\small
\caption{Greedy final-answer accuracy (\%) on Qwen3-4B with alternative distillation objectives. All distillation runs use student prefixes.}
\label{tab:v3_variants}
\setlength{\tabcolsep}{8pt}
\begin{tabular}{@{}lcccc@{}}
\toprule
& \multicolumn{2}{c}{Teacher $T_f$} & \multicolumn{2}{c}{Teacher $T_s$} \\
\cmidrule(lr){2-3}\cmidrule(lr){4-5}
Objective & \textsc{Fact-Single} & \textsc{Comp-Unseen} & \textsc{Fact-Single} & \textsc{Comp-Unseen} \\
\midrule
$S_0$ (no distillation) & 0.1 & 1.8 & 0.1 & 1.8 \\
\midrule
Reverse KL & 10.9 & 0.7 & 0.1 & 76.8 \\
Forward KL & 54.1 & 0.5 & 0.1 & 76.4 \\
GKD ($\beta=0.5$) & 9.1 & 0.7 & 0.1 & 70.6 \\
ToDi & 64.3 & 0.5 & 0.1 & 86.4 \\
EOPD & 13.7 & 0.7 & 0.1 & 78.7 \\
\bottomrule
\end{tabular}
\end{table}

Table~\ref{tab:v3_variants} shows that ToDi achieves the highest accuracy on both target tests: 64.3\% factual recall after distillation from $T_f$ and 86.4\% on unseen tree structures after distillation from $T_s$. These exceed forward KL by 10.2 and 10.0 percentage points, respectively. GKD and EOPD retain the transfer asymmetry, reaching 70.6\% and 78.7\% on unseen tree structures but only 9.1\% and 13.7\% factual recall. Thus, hybrid divergence objectives can improve both factual and compositional transfer, with performance varying substantially across objectives. Across all objectives, distillation from $T_f$ leaves compositional accuracy near zero, while distillation from $T_s$ leaves held-out factual recall at its initial level.

\section{Conclusion}

This paper investigates what language models acquire through on-policy distillation by separating factual knowledge  from multi-step compositional skill. Across synthetic tasks and real-world benchmarks, we observe a consistent transfer asymmetry:  standard reverse-KL OPD transfers compositional skill strongly but factual knowledge only weakly. By separating the source of training prefixes from the divergence objective, we show that factual transfer is primarily determined by the KL objective, whereas student rollouts improve multi-step compositional execution. Forward KL substantially improves factual transfer under both student- and teacher-generated prefixes. Overall, these findings suggest that standard on-policy distillation is better suited for improving how language models organize and use existing knowledge than for updating factual knowledge. When factual knowledge transfer is the goal, forward-KL supervision provides a more effective alternative. 

\bibliographystyle{plainnat}
\bibliography{refs}


\clearpage
\appendix

\section*{Appendix}
\pdfbookmark[0]{Appendix}{appendix-overview}
\etocdepthtag.toc{appendix}
\etocsettagdepth{main}{none}
\etocsettagdepth{appendix}{subsection}
\etocsettocstyle{\subsection*{Contents}}{}
\begingroup
\hypersetup{linkcolor=black}
\tableofcontents
\endgroup
\clearpage

\section{Execution Examples}
\label{sec:execution_examples}

We provide complete input prompts and reference execution traces for a chain and a branching tree. Both examples use the same eight four-digit inputs and query each of the same seven initial facts, identified by \texttt{[OP\_1]} through \texttt{[OP\_7]}, once. The structures differ in how intermediate results are combined. All values are computed from the experimental lookup tables. Line breaks and indentation are added for readability. Each completion consists of semicolon-separated steps followed by the final answer within \texttt{[ANS]} and \texttt{[/ANS]}.

\paragraph{Sequential Chain}
Computation proceeds from left to right. Each step after the first uses the previous step's output as its left operand and the next leaf input as its right operand.

\begin{quote}
\small
\textit{Input prompt}\par\smallskip
{\ttfamily\noindent
((((((1234 [OP\_1] 5678) [OP\_2] 9012) [OP\_3] 3456)\newline
\hspace*{1em}[OP\_4] 7890) [OP\_5] 2468) [OP\_6] 1357) [OP\_7] 8024 = ?\par}
\medskip
\textit{Reference completion}\par\smallskip
{\ttfamily\noindent
1234[OP\_1]5678=8495;\newline
8495[OP\_2]9012=7688;\newline
7688[OP\_3]3456=7701;\newline
7701[OP\_4]7890=2436;\newline
2436[OP\_5]2468=8325;\newline
8325[OP\_6]1357=6416;\newline
6416[OP\_7]8024=9196;\newline
[ANS] 9196 [/ANS]\par}
\end{quote}

\paragraph{Branching Tree}
Following left-to-right postorder traversal, steps 1 to 3 evaluate the left subtree, steps 4 to 6 evaluate the right subtree, and step 7 combines their results.

\begin{quote}
\small
\textit{Input prompt}\par\smallskip
{\ttfamily\noindent
((1234 [OP\_1] 5678) [OP\_3] (9012 [OP\_2] 3456)) [OP\_7]\newline
\hspace*{1em}((7890 [OP\_4] 2468) [OP\_6] (1357 [OP\_5] 8024)) = ?\par}
\medskip
\textit{Reference completion}\par\smallskip
{\ttfamily\noindent
1234[OP\_1]5678=8495;\newline
9012[OP\_2]3456=9508;\newline
8495[OP\_3]9508=7122;\newline
7890[OP\_4]2468=3507;\newline
1357[OP\_5]8024=5086;\newline
3507[OP\_6]5086=8064;\newline
7122[OP\_7]8064=3761;\newline
[ANS] 3761 [/ANS]\par}
\end{quote}

At step 7, the left operand \texttt{7122} comes from step 3, while the right operand \texttt{8064} comes from step 6. The model must retrieve the left subtree's result after three intervening steps on the right subtree. This balanced structure belongs to the unseen pool defined in Appendix~\ref{sec:shape_splits}.

\section{Additional Experimental Details}
\label{sec:v31_details}

\subsection{Lookup Tables and Data Construction}
Table entries are sampled uniformly from the ten digits. Values are represented as four-digit strings with leading zeros. The tokenizer adds 25 fact identifiers, the identity marker \texttt{[OP\_ID]}, and two answer delimiters. The identity marker is excluded from the initial/held-out fact partition.

\subsection{Training Corpora and Budgets}
\label{sec:training_corpora}
Each student-initialization or teacher-training corpus has an estimated token budget of 10 million. The estimate counts each special token and each remaining non-whitespace character as one unit, summed over prompts and reference completions.

The $S_0$ corpus allocates 45\% of this budget to single-step queries about initial facts, 25\% to chains of two, three, five, or seven steps, and 30\% to groups of seven independent single-step queries, totaling 166,365 records. Each group is presented in one prompt; its queries use separate inputs and do not depend on one another's results. The completion lists the seven single-step computations and encloses their comma-separated answers within \texttt{[ANS]} and \texttt{[/ANS]}.

Each trained teacher allocates 40\% of its budget to the initialization tasks. Within this portion, 30\% covers single-step queries, 40\% chains, and 30\% groups of seven independent queries, all over initial facts. The remaining 60\% differs by teacher. For $T_f$, 30\% of the total budget covers single-step queries about held-out facts and 30\% covers chains querying held-out facts. For $T_s$, 60\% covers trees over initial facts. For $T_{fs}$, 20\% covers single-step queries about held-out facts, 20\% trees over initial facts, and 10\% each chains and trees querying held-out facts. Single-step queries about held-out facts are split equally by budget between individual queries and groups of seven. All teacher-training chains and trees and all distillation data contain seven steps.

Each distillation corpus contains 20,000 prompts, with queries excluded from student initialization and teacher training. In teacher training and distillation, chains and trees querying held-out facts contain one, three, or seven such steps, sampled with probabilities 0.5, 0.3, and 0.2, respectively; any remaining steps query initial facts.

\subsection{Tree Structures, Test Sets, and Metrics}
\label{sec:tests_metrics}

\paragraph{Tree Structures Pools}
\label{sec:shape_splits}
Of the 429 ordered full binary-tree structure with seven internal nodes, 64 form the shared pool, 64 the teacher-only pool, and 128 the unseen pool; 173 are unused. The balanced tree of height three, where height counts edges on the longest root-to-leaf path, belongs to the unseen pool. The teacher-only pool includes the left-associative chain used during student initialization. Thus, teacher-only denotes exclusion from distillation prompts, not complete exclusion from the student's earlier training.

\paragraph{Test Set Definitions}
The factual knowledge tests use held-out facts. The compositional skill tests use only initial facts and differ in their tree structure pool. Each test contains 2,000 queries excluded from the training corpora. All multi-step tests use seven computational steps.

\begin{center}
\small
\begin{tabularx}{\textwidth}{@{}l>{\raggedright\arraybackslash\leavevmode}X@{}}
\toprule
Test set & Definition \\
\midrule
\textsc{Fact-Single} & Single-step queries, each querying one held-out fact. \\
\textsc{Fact-Chain} & Chains in which all steps query held-out facts. Fact identifiers may repeat. \\
\textsc{Comp-Shared} & Trees over initial facts with structures from the shared pool. \\
\textsc{Comp-TeacherOnly} & Trees over initial facts with structures from the teacher-only pool. \\
\textsc{Comp-Unseen} & Trees over initial facts with structures from the unseen pool. \\
\bottomrule
\end{tabularx}
\end{center}

\paragraph{Evaluation Metrics}
\emph{Greedy accuracy} is the percentage of queries whose greedily decoded answer exactly matches the reference four-digit answer, including leading zeros. Answers are extracted between \texttt{[ANS]} and \texttt{[/ANS]}; missing or unparseable answers count as incorrect. This is the metric reported in Table~\ref{tab:teacher_profiles_eval}.

\emph{Pass@32} is the percentage of queries for which at least one of 32 independently sampled completions has the correct final answer, with sampling temperature 1.0. It is distinct from greedy final-answer accuracy and does not require the entire execution trace to be correct.

\subsection{Student and Teacher Baselines}
Table~\ref{tab:teacher_profiles_eval} reports greedy final-answer accuracy for $S_0$ and the specialized teachers across four models from three families: Qwen3-0.6B and Qwen3-4B~\citep{2025qwen3}, Gemma-2-2B~\citep{team2024gemma}, and OLMo3-7B~\citep{olmo2025olmo}.

For Qwen3-4B, $S_0$ achieves 100\% greedy final-answer accuracy on seven-step chains over initial facts, the chain task used during initialization, compared with 0.45\% on \textsc{Comp-Shared}. Its final-answer Pass@32 on \textsc{Comp-Shared} is 1.45\%.

\begin{table}[htbp]
  \centering
  \caption{Greedy final-answer accuracy (\%) of $S_0$ and the specialized teachers on factual knowledge and compositional skill tests. Test definitions appear in Appendix~\ref{sec:tests_metrics}. All tests use exact match of the final four-digit answer. Green shading highlights high accuracy.}
  \label{tab:teacher_profiles_eval}
  \footnotesize
  \setlength{\tabcolsep}{5pt}
  \renewcommand{\arraystretch}{1.1}
  \definecolor{qgreen}{RGB}{46, 139, 87}
  \providecommand{\hi}{\cellcolor{qgreen!18!white}}   
  \providecommand{\md}{\cellcolor{qgreen!12!white}}   
  \providecommand{\lo}{\cellcolor{qgreen!8!white}}   
  \begin{tabular}{@{}llccccc@{}}
    \toprule
    & & \multicolumn{2}{c}{\textbf{Factual Knowledge}} & \multicolumn{3}{c}{\textbf{Compositional Skill}} \\
    \cmidrule(lr){3-4}
    \cmidrule(lr){5-7}
    \textbf{Model} & \textbf{Profile}
      & \textsc{Fact-Single} & \textsc{Fact-Chain}
      & \shortstack{\textsc{Comp-}\\\textsc{Shared}} & \shortstack{\textsc{Comp-}\\\textsc{TeacherOnly}} & \shortstack{\textsc{Comp-}\\\textsc{Unseen}} \\
    \midrule
    \multirow{4}{*}{\textbf{Qwen3-0.6B}}
     & $S_0$ (student) & 0.00 & 0.00 & 0.15 & 2.10 & 0.20 \\
     & $T_f$           & \hi 100 & \hi 100 & 0.20 & 2.05 & 0.45 \\
     & $T_s$           & 0.00 & 0.00 & \hi 100 & \hi 100 & \hi 99.95 \\
     & $T_{fs}$        & \hi 100 & \hi 100 & \hi 100 & \hi 99.95 & \hi 100 \\
    \midrule
    \multirow{4}{*}{\textbf{Gemma-2-2B}}                                                                                 & $S_0$ (student) & 0.00 & 0.00 & 0.40 & 2.20 & 0.45 \\                                                            & $T_f$           & \hi 100 & \hi 100 & 0.30 & 2.20 & 0.10 \\                                                       & $T_s$           & 0.00 & 0.05 & \hi 100 & \hi 100 & \hi 99.95 \\                                                 & $T_{fs}$        & \hi 100 & \hi 99.95 & \hi 100 & \hi 99.95 & \md 99.55 \\                                      \midrule
        \multirow{4}{*}{\textbf{Qwen3-4B}}
     & $S_0$ (student) & 0.10 & 0.00 & 0.45 & 2.75 & 1.80 \\
     & $T_f$           & \hi 100 & \hi 99.95 & 0.25 & 2.30 & 0.60 \\
     & $T_s$           & 0.00 & 0.00 & \hi 100 & \hi 100 & \hi 100 \\
     & $T_{fs}$        & \hi 100 & \hi 99.95 & \hi 100 & \hi 100 & \hi 100 \\
    \midrule
    \multirow{4}{*}{\textbf{OLMo3-7B}}
     & $S_0$ (student) & 0.00 & 0.00 & 0.35 & 2.20 & 0.45 \\
     & $T_f$           & \hi 100 & \hi 99.95 & 0.35 & 2.25 & 0.25 \\
     & $T_s$           & 0.00 & 0.00 & \md 99.10 & \md 99.45 & \lo 97.80 \\
     & $T_{fs}$        & \hi 100 & \md 99.50 & \md 99.05 & \md 99.00 & \md 98.85 \\
    \bottomrule
  \end{tabular}
\end{table}

\subsection{Optimization and Hyperparameters}
All experiments use initialization and training seed 1. Student initialization and teacher training use completion cross-entropy with a maximum sequence length of 1,024 tokens, a learning rate of $10^{-4}$, and three epochs. OPD uses a learning rate of $3\times10^{-6}$ and trains for three epochs on 20,000 prompts per distillation condition. Each visit to a prompt generates one student rollout at temperature 1.0, capped at 256 new tokens. The reverse KL is computed over the full vocabulary.

All training stages employ AdamW with a cosine learning-rate schedule, 3\% linear warmup, zero weight decay, gradient norm clipping at 1.0, and bfloat16 precision. The global batch size is 32 for the Qwen3 and Gemma-2 models and 16 for OLMo3-7B. Student initialization, teacher training, and OPD use LoRA adapters~\citep{hu2022lora} on the attention and feed-forward projection layers (rank 16, $\alpha=32$, dropout 0.05). Teachers and distillation students initialize from $S_0$ with its adapters merged into the base weights, then train fresh adapters. Gradient masking restricts token-embedding updates to newly added token rows.

\section{Training Dynamics}
\label{sec:training_dynamics}

\begin{figure}[h]
\centering
\includegraphics[width=\textwidth]{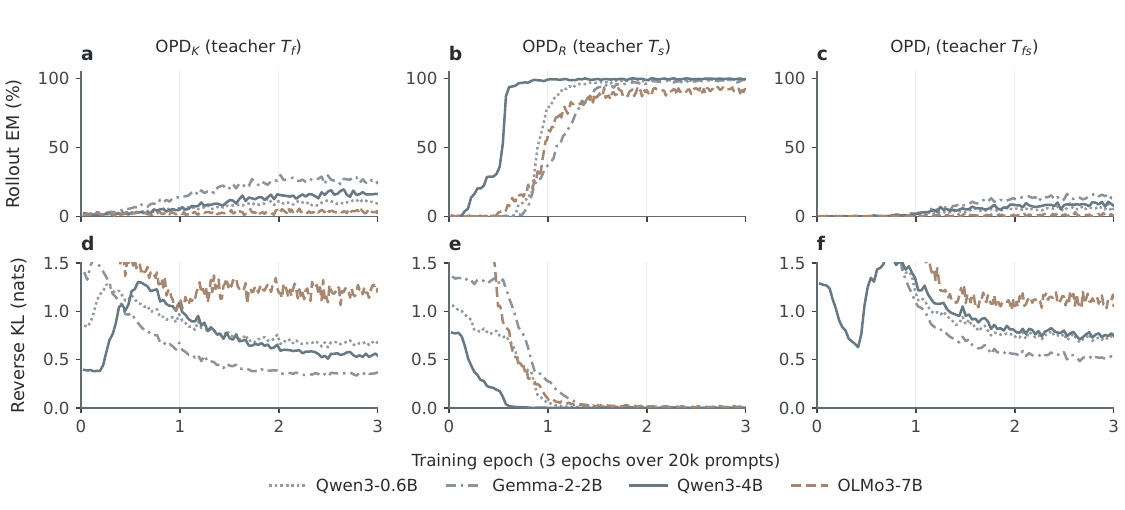}
\caption{Training curves for on-policy distillation in the synthetic experiments. Columns correspond to distillation from $T_f$ (a, d), $T_s$ (b, e), and the joint $T_{fs}$ (c, f). Top row: exact-match accuracy of sampled student rollouts on the distillation prompts; bottom row: reverse KL in nats. All runs train for three epochs over 20,000 prompts.}
\label{fig:training_curves}
\end{figure}

Figure~\ref{fig:training_curves} shows how rollout accuracy and the reverse-KL objective evolve over the three training epochs. Under $T_s$, rollout exact match rises from near zero to above 95\% in all four models, and the reverse KL falls toward zero: once the student produces correct executions, its own trajectories coincide with the teacher's supervision. Under $T_f$, rollout exact match does not exceed about 30\% in any model, and the reverse KL plateaus above zero instead of vanishing. The residual gap concentrates at positions whose answer depends on a held-out fact, where the teacher is confident about the correct digits while the student spreads probability across candidates it cannot distinguish. The joint $T_{fs}$ run shows the same plateau.

\section{Real-World Benchmarks and Teacher Profiles}
\label{sec:real_world_details}

To evaluate whether the transfer asymmetry observed in synthetic tasks extends to natural language, we curate two evaluation suites covering post-cutoff factual knowledge and competition-level mathematical reasoning.

\paragraph{Benchmark Construction}
The \textsc{Knowledge-2026} benchmark evaluates factual recall using 809 question-answer pairs that entered public knowledge in 2026, combining items from RealTimeQA~\citep{kasai2023realtime}, SealQA~\citep{pham2026sealqa}, FreshQA~\citep{vu2024freshllms}, and newly edited Wikipedia articles extracted via the WikiBigEdit pipeline~\citep{pmlr-v267-thede25a}. Because these items postdate the pretraining cutoff of Qwen3-4B-Base, the base student cannot rely on memorized pretraining associations. The \textsc{Math-2026} benchmark evaluates multi-step reasoning using 201 competition-level math problems with verified step-by-step solutions, sampled from MathArena~\citep{balunovic2026matharena} and NuminaMath-CoT~\citep{numina_math_datasets} and held out from the curated 1,000-problem pool described in Section~\ref{sec:real_world}.

\paragraph{Teacher Profiles}
The initial student $S_0$ is fine-tuned from Qwen3-4B-Base using Tulu-3-SFT data~\citep{lambert2024tulu}. The specialized teachers share $S_0$'s initialization and training hyperparameters, but incorporate domain-specific supervision: $T_K$ is fine-tuned on the full set of 809 2026 factual QA pairs, and $T_R$ is fine-tuned on verified competition-math solutions. 


\section{Weight Update Measurements}
\label{sec:adapter_metrics}
We evaluate parameter changes by analyzing the merged LoRA weight increments relative to $S_0$. For each adapted projection $m$, the increment is $\Delta W_m = (\alpha/r) B_m A_m \in \mathbb{R}^{d_{\mathrm{out}} \times d_{\mathrm{in}}}$, with rank $r=16$ and scaling factor $\alpha=32$. Below we define the metrics reported in Table~\ref{tab:adapter_anatomy} and Table~\ref{tab:adapter_anatomy_full}.

\paragraph{Update Norms}
The global update norm measures the aggregate magnitude of parameter changes across all adapted linear projections, excluding embeddings:
\begin{equation}
    \|\Delta W\|_F = \sqrt{\sum_m \|\Delta W_m\|_F^2}.
\end{equation}
To inspect depth and architectural localization, we compute sub-norms by restricting the sum to specific modules. The layer norm $\|\Delta W_\ell\|_F$ aggregates all projections in layer $\ell$, while the projection norm $\|\Delta W_p\|_F$ aggregates projection type $p \in \{q, k, v, o, \text{gate}, \text{up}, \text{down}\}$ across all layers. For depth comparisons in Qwen3-4B, layer norms are averaged within four consecutive depth quartiles of nine layers each (Layers 0--8, 9--17, 18--26, and 27--35).

\paragraph{Directional Similarity}
To measure alignment between two training conditions within the same model family, we concatenate their projection increments into a single vector $\mathbf{v} = \mathrm{vec}(\Delta W)$. The cosine similarity between conditions (1) and (2) is defined as
\begin{equation}
    \cos(\mathbf{v}^{(1)}, \mathbf{v}^{(2)}) = \frac{\langle \mathbf{v}^{(1)}, \mathbf{v}^{(2)} \rangle}{\|\mathbf{v}^{(1)}\|_2 \|\mathbf{v}^{(2)}\|_2}.
\end{equation}

\paragraph{Effective Rank}
For each projection increment $\Delta W_m$ with singular values $\sigma_{m,1} \ge \dots \ge \sigma_{m,r} > 0$, we define the normalized spectral distribution $p_{m,i} = \sigma_{m,i}^2 / \sum_j \sigma_{m,j}^2$. The effective rank is the exponential Shannon entropy of this distribution,
\begin{equation}
    r_{\mathrm{eff}}(\Delta W_m) = \exp\left(-\sum_{i=1}^r p_{m,i} \log p_{m,i}\right),
\end{equation}
averaged across all adapted projections. This metric reflects how uniformly update energy is distributed across singular directions, ranging from 1 for a rank-1 update to 16 for a uniform spectrum.

\paragraph{Embedding Drift}
Embedding drift measures the Euclidean displacement of fact-identifier tokens from their $S_0$ initialization, $\|\mathbf{e}_k - \mathbf{e}_k^{(S_0)}\|_2$. We report the mean drift separately for held-out tokens (\texttt{[OP\_21]} through \texttt{[OP\_25]}) and a representative sample of initial tokens (\texttt{[OP\_1]} through \texttt{[OP\_5]}).

\begin{table}[htbp]
\centering
\footnotesize
\caption{Complete weight updates relative to $S_0$. (a) Global update norm / mean effective rank. (b) Cosine similarity between updates. (c) Mean embedding drift for held-out / initial fact identifiers. (d) Mean layer norms by depth quartile. (e) Norms by projection. Ranges cover the four prefix--KL combinations; control ranges cover two prompt sets.}
\label{tab:adapter_anatomy_full}
\setlength{\tabcolsep}{7pt}
\renewcommand{\arraystretch}{1.05}
\begin{tabular}{@{}lccc@{}}
\toprule
\textbf{(a) Teacher: prefix, objective} & Qwen3-4B & Gemma-2-2B & OLMo3-7B \\
\midrule
$T_f$: student, reverse KL & 2.01 / 2.4 & 1.33 / 3.9 & 3.08 / 2.3 \\
$T_f$: student, forward KL & 2.64 / 2.5 & 1.49 / 3.6 & 3.85 / 2.0 \\
$T_f$: teacher, forward KL & 2.86 / 2.5 & 1.52 / 3.6 & 3.89 / 2.1 \\
$T_f$: teacher, reverse KL & 2.27 / 2.5 & 1.34 / 3.8 & 3.20 / 2.1 \\
$T_s$: student, reverse KL & 1.27 / 3.3 & 1.13 / 3.9 & 2.50 / 2.6 \\
$T_s$: student, forward KL & 1.55 / 2.7 & 1.60 / 3.6 & 3.21 / 2.5 \\
$T_s$: teacher, forward KL & 1.52 / 3.1 & 1.45 / 4.0 & 3.06 / 2.9 \\
$T_s$: teacher, reverse KL & 1.42 / 3.3 & 1.27 / 4.0 & 2.68 / 3.0 \\
$T_\emptyset$: factual prompts & 0.30 / 10.2 & 0.20 / 12.7 & 0.48 / 11.7 \\
$T_\emptyset$: compositional prompts & 0.32 / 10.8 & 0.24 / 12.7 & 0.55 / 12.1 \\
\midrule
\multicolumn{4}{@{}l}{\textbf{(b) Update cosine similarity}} \\
\midrule
$T_f$ vs. $T_s$ (matched prefix and KL) & 0.004--0.018 & 0.011--0.036 & 0.004--0.046 \\
$T_f$: reverse vs. forward KL, student & 0.53 & 0.49 & 0.48 \\
$T_f$: reverse vs. forward KL, teacher & 0.50 & 0.50 & 0.50 \\
$T_s$: reverse vs. forward KL, student & 0.73 & 0.64 & 0.68 \\
$T_s$: reverse vs. forward KL, teacher & 0.78 & 0.82 & 0.80 \\
\midrule
\multicolumn{4}{@{}l}{\textbf{(c) Embedding drift: held-out / initial identifiers}} \\
\midrule
$T_f$ (supervised fine-tuning) & 0.0420 / 0.0630 & 0.0290 / 0.0910 & 0.0840 / 0.1570 \\
Distilled students (maximum across conditions) & 0.0071 / 0.0060 & 0.0072 / 0.0020 & 0.0024 / 0.0021 \\
$T_\emptyset$ controls (maximum across prompt sets) & 0.0004 / 0.0005 & 0.0002 / 0.0004 & 0.0005 / 0.0007 \\
\bottomrule
\end{tabular}
\par\smallskip
\begin{tabular}{@{}lccccc@{}}
\toprule
\textbf{(d) Qwen3-4B: layer norms} & Layers 0--8 & 9--17 & 18--26 & 27--35 & Peak layer \\
\midrule
$T_f$ & 0.24--0.33 & 0.25--0.39 & 0.33--0.46 & 0.46--0.65 & 35 \\
$T_s$ & 0.25--0.29 & 0.26--0.27 & 0.18--0.25 & 0.11--0.24 & 3, 11, 34 \\
$T_\emptyset$ & 0.04--0.05 & 0.05--0.05 & 0.05--0.05 & 0.06--0.06 & 29 \\
\bottomrule
\end{tabular}
\par\smallskip
\begin{tabular}{@{}lccccccc@{}}
\toprule
\textbf{(e) Qwen3-4B: teacher, forward KL} & $q$ & $k$ & $v$ & $o$ & gate & up & down \\
\midrule
$T_f$ & 0.95 & 0.45 & 0.40 & 0.74 & 1.76 & 1.61 & 0.81 \\
$T_s$ & 0.50 & 0.24 & 0.26 & 0.49 & 0.82 & 0.85 & 0.55 \\
\bottomrule
\end{tabular}
\end{table}
\end{document}